%% file: main.tex
\documentclass[journal,twoside,web]{ieeecolor}

\usepackage{etoolbox}
\makeatletter
\@ifundefined{color@begingroup}%
  {\let\color@begingroup\relax
   \let\color@endgroup\relax}{}%
\def\fix@ieeecolor@hbox#1{%
  \hbox{\color@begingroup#1\color@endgroup}}
\patchcmd\@makecaption{\hbox}{\fix@ieeecolor@hbox}{}{\FAILED}
\patchcmd\@makecaption{\hbox}{\fix@ieeecolor@hbox}{}{\FAILED}

\usepackage{lcsys}
\usepackage{cite}

\usepackage{amsmath,amssymb,amsfonts}
\usepackage{cleveref}
\usepackage{algorithm}
\usepackage{algorithmic}
\usepackage{graphicx}
\usepackage{textcomp}
\usepackage{fancyhdr}
\newcommand{\diag}{\operatorname{diag}}
\newcommand{\tr}{\mathsf{ T}}
\newcommand{\poly}{\operatorname{poly}}

\newtheorem{assumption}{Assumption}
\crefname{assumption}{Assumption}{Assumptions}
\newtheorem{lemma}{Lemma}
\crefname{lemma}{Lemma}{Lemmas}

\newtheorem{theorem}{Theorem}
\crefname{theorem}{Theorem}{Theorems}

\usepackage{siunitx}
\usepackage{threeparttable}
\usepackage{booktabs}
\makeatletter
\def\ps@IEEEtitlepagestyle{%
  \def\@oddhead{}%
  \def\@evenhead{}%
  \def\@oddfoot{}%
  \def\@evenfoot{}%
}
\makeatother

\def\BibTeX{{\rm B\kern-.05em{\sc i\kern-.025em b}\kern-.08em
    T\kern-.1667em\lower.7ex\hbox{E}\kern-.125emX}}

\begin{document}

\title{
Logarithmic Regret and Polynomial Scaling in Online Multi-step-ahead Prediction
}

\author{Jiachen~Qian,~\IEEEmembership{Student Member,~IEEE,}
       Yang Zheng,~\IEEEmembership{Senior Member,~IEEE}
\thanks{This work is supported by NSF CMMI 2320697 and NSF CAREER 2340713. Jiachen Qian and Yang Zheng are with the Department of Electrical and
Computer Engineering, University of California San Diego, 
CA 92093 USA (e-mail: jiq012@ucsd.edu; zhengy@ucsd.edu)}}

\maketitle

\thispagestyle{plain}
\begin{abstract}
This letter studies the problem of online multi-step-ahead prediction for unknown linear stochastic systems.
Using conditional distribution theory, we derive an optimal parameterization of the prediction policy as a linear function of future inputs, past inputs, and past outputs. Based on this characterization, we propose an online least-squares algorithm to learn the policy and analyze its regret relative to the optimal model-based predictor. We show that the online algorithm achieves logarithmic regret with respect to the optimal Kalman filter in the multi-step setting. Furthermore, with new proof techniques, we establish an almost-sure regret bound that does not rely on fixed failure probabilities for sufficiently large horizons $N$. Finally, our analysis also reveals that, while the regret remains logarithmic in $N$, its constant factor grows polynomially with the prediction horizon $H$, with the polynomial order set by the largest Jordan block of eigenvalue 1 in the system matrix.
\end{abstract}

\begin{IEEEkeywords}
Model-free learning, Multi-step prediction, Logarithmic Regret
\end{IEEEkeywords}

\input{Section-Introduction}
\input{Section-Preliminary}
\input{Section-MainResult}
\input{Section-Simulation}

\section{Conclusion}
In this letter, we address the problem of multi-step-ahead prediction for unknown linear stochastic systems. We have derived an optimal parameterization of the $H$-step predictor as a linear combination of future inputs, past inputs, and past outputs. We have proposed an online least–squares-based algorithm to learn this policy. Our algorithm achieves an almost-sure logarithmic regret bound with respect to the optimal model-based $H$-step predictor. The dependence on the time horizon $N$ is logarithmic, while the multiplicative constant scales polynomially with the prediction horizon $H$. 
Future directions include {\color{black}extensions to structured nonlinear systems and} designing online feedback policies to stabilize a linear stochastic system with regret guarantees relative to an optimal LQR/LQG controller.

\bibliographystyle{IEEEtran}
\bibliography{ref}
 
\appendix
\input{Section-Proof}

\end{document}

%% file: Section-Introduction.tex
\section{Introduction}
 \IEEEPARstart{O}{nline} 
 prediction of dynamical system behavior has long been recognized as a fundamental problem in  
 control systems \cite{Andersonoptimal}, robotics \cite{barfoot2024state}, computer vision \cite{coskun2017long}, etc. 
 Classical approaches rely on an accurate model and known noise statistics to propagate system responses and examine the effect of future control inputs
 \cite{kailath2000linear}. For instance, the celebrated Kalman filter provides 
 \textit{optimal} mean-square-error predictions under correct modeling assumptions
 \cite{kalmanfilter}. 
 In many applications \cite{li2017dynamical,wu2019machine,markovsky2023data}, however, obtaining explicit models and reliable noise characterizations is impractical, especially when input-output relationships are complex to identify. This has motivated growing interest in learning prediction policies directly from data, without full knowledge of the underlying system.

From model-based prediction to data-driven approaches,~a central challenge is how to parameterize the prediction policy. Traditional system identification addresses this by first identifying Markov parameters and then solving a nonconvex problem to extract a system model, upon which a standard Kalman predictor can be used \cite{ljung1998system}. Recent advances~have~provided non-asymptotic analysis of the identification process by characterizing convergence rates of the identification error  \cite{ziemann2023tutorial, he2025finite} and leveraging multiple trajectories to mitigate potential state divergence \cite{zheng2020non}. Nonetheless, mapping from Markov parameters to a system model is intrinsically nonlinear and nonconvex \cite{hardt2018gradient}, which makes it difficult to establish strong theoretical guarantees for online prediction performance.

Instead of identifying an explicit model, some recent~studies learn a prediction policy directly from input-output data \cite{hazan2017learning, tsiamis9894660, rashidinejad2020slip, qian2025model, zhang2023learningTK}. These methods exploit the Kalman filter's structure and parameterize the prediction policy as a linear function of past inputs and outputs, and then estimate the~weights via online learning. This approach, also known as {\it improper learning}, bypasses the intermediate system identification. Notably, \cite{tsiamis9894660} shows that by employing a truncated autoregressive (AR) model derived from the Kalman filter, a regret measure (i.e., the cumulative loss of the online predictor relative to the optimal model-based prediction) scales logarithmically with the time horizon. Thus, with sufficiently long trajectories, the average prediction error of the online predictor approaches that of the optimal Kalman filter. Building on this, \cite{rashidinejad2020slip} established a similar logarithmic regret bound using low-rank approximation techniques to address slow convergence under heavy noise. 
Our recent work \cite{qian2025model} has introduced~an \textit{exponential forgetting strategy} to address the unbalanced regression model in online prediction while preserving the logarithmic~regret.

The aforementioned results \cite{hazan2017learning, tsiamis9894660, rashidinejad2020slip, qian2025model, zhang2023learningTK} focus primarily on \textit{single-step} prediction. In practical scenarios such as path~planning and predictive control  \cite{tsiamis2024predictive,wang2023deep,liu2017path}, \textit{multi-step} predictions are essential to enforce state~constraints and optimize control policies. While empirical studies \cite{farina2008some,lambert2021learning} suggest that recursively applying a single-step predictor can extend the prediction horizon, such autoregressive roll-outs can compound errors and often underperform direct multi-step prediction \cite{somalwar2025learning}. 
Yet, the theory of effective \textit{multi-step} prediction,  especially a quantitative characterization of performance degradation with horizon length, remains largely underdeveloped. 

In this paper, we focus on model-free online learning of a \textit{multi-step-ahead} predictor for linear stochastic systems.
Our contributions are as follows. 
  First,  based on conditional distribution theory, we introduce an autoregressive model that parameterizes the $H$-step-ahead prediction policy as a linear combination of past outputs, past inputs, and future outputs (\Cref{proposition:innovation}). 
  Unlike single-step prediction, the innovation in $H$-step-ahead prediction becomes temporally correlated and \textit{non-orthogonal}. We further establish that the innovation term coincides with the autoregressive roll-out in \cite{tsiamis9894660}, thereby providing a theoretical justification for the heuristic approach. 
   Second, based on the autoregressive model, we propose an online least-squares-based learning algorithm for $H$-step-ahead prediction. {With a backward horizon chosen proportional to $\log N$, where $N$ denotes the total time horizon, we establish that the regret with respect to the optimal model-based Kalman predictor scales logarithmically in $N$ almost surely (\cref{thm1}).} The regret remains logarithmic despite the non-orthogonal innovation process. {The prediction horizon $H$ does not change the order in $N$; instead, it appears only in the constant, which grows polynomially at a rate no larger than $H^{4\kappa+1}$, where $\kappa$ is the size of the largest Jordan block of eigenvalue $1$ in the system matrix.}
   Compared with the prior literature \cite{hazan2017learning, tsiamis9894660, rashidinejad2020slip, zhang2023learningTK, qian2025model}, our results give the first explicit regret scaling in terms of prediction horizon and hold almost surely rather than merely in probability.

{\it Notation:} We use $A\succ B$ to denote that $A - B$ is positive definite. We use $\left\|\cdot\right\|_2$, $\left\|\cdot\right\|_F$, and $\left\|\cdot\right\|_1$ to  denote the 2-norm, the Frobenius norm, and  the 1-norm, respectively. $\mathcal{N}(\mu, V)$ denotes a Gaussian distribution with mean $\mu$ and variance~$V$. 
$\rho(A)$ denotes the spectral radius of $A$. $\operatorname{poly}(x)$ denotes a polynomial in $x$. $\mathcal{O}(f(x))$ indicates the function is of the same order as $f(x)$, $o(f(x))$ indicates the function is of a smaller order than $f(x)$.

%% file: Section-Preliminary.tex
\section{Preliminaries and problem statement}
\subsection{Linear stochastic system and the Kalman filter}
Consider the following linear stochastic system
\begin{equation}\label{eq: LinearSystem}
\begin{aligned}
x_{k+1}&=\;Ax_k +Bu_k +w_k,\\ y_{k}&=\;Cx_k + v_{k}, \quad k = 0,1,2,\ldots 
\end{aligned}
\end{equation}
where $x_k\in\mathbb{R}^n$ is the state vector, $u_k\in\mathbb{R}^{n_u}$ is the input vector, $y_{k}\in\mathbb{R}^{m}$ is the output vector, $\omega_k\sim \mathcal{N}(0,Q)$ and  $v_{k}\sim \mathcal{N}(0,R)$ are the process and observation noises with $Q,R\succ 0$, respectively, and we assume $u_k\sim\mathcal{N}(0,I_{n_u})$.

In this paper, we make a standard assumption.
\begin{assumption}\label{asp: System}
    The matrix $A$ is marginally stable, i.e., $\rho(A)\leq 1$, and the system pair $(A, C)$ is detectable.
\end{assumption}

If the system parameters $(A,B,C,Q,R)$ are known, we can apply the Kalman filter \cite{kalmanfilter,Andersonoptimal} to predict the future outputs. 
Let  $\mathcal{F}_{k} \triangleq \sigma\left(y_{0}, \ldots, y_{k}\right)$  be the filtration generated by the observations  $y_{0}, \ldots, y_{k}$, \textcolor{black}{where $\sigma$ is the sigma algebra}. Given $\mathcal{F}_{k}$, we aim to predict the optimal $\hat{y}_{k+1}$ in the minimum mean-square error~sense: 
$
    \hat{y}_{k+1} \triangleq \arg \min _{z \in \mathcal{F}_{k}} \mathbb{E}\left[\left\|y_{k+1}-z\right\|_{2}^{2} \mid \mathcal{F}_{k}\right].
$
It is now well-known that the steady-state optimal predictor takes a recursive form, known as the \textit{Kalman filter} \cite{kalmanfilter}, 
\begin{equation}\label{eq: KalmanPredictor}
\begin{aligned}
\hat{x}_{k+1} & =A \hat{x}_{k}+Bu_k+L\left( y_{k}-\hat{y}_{k}\right) 
,\;\;\hat{x}_{0}=0 \\
\hat{y}_{k} & =C \hat{x}_{k}, 
\end{aligned}
\end{equation}
where $L\!=\!APC^\tr\left(CPC^\tr\!+\!R\right)^{-1}$ is called the steady-state Kalman gain with \textcolor{black}{$P$ from 
the algebraic Riccati equation: 
\begin{equation}\label{eq: DARE}
    P=APA^\tr+Q-APC^\tr\left(CPC^\tr+R\right)^{-1}CPA^\tr, 
\end{equation}
which has a unique positive definite stabilizing solution with \cref{asp: System} and $Q,R\succ 0$ \cite{kailath2000linear}.}
It is also shown that the optimal prediction $\hat{y}_{k+1}$ is equivalent to the expectation of $y_{k+1}$ conditioned on $\mathcal{F}_k$, i.e., $\hat{y}_{k+1}=\mathbb{E}\left\{y_{k+1}\mid \mathcal{F}_k\right\}$ \cite[Section 2]{Andersonoptimal}. 
We note that the general Kalman filter takes a time-varying form. However, due to the exponential convergence of the conditioned output process $y_k$ to become steady-state \cite{Andersonoptimal}, the difference between the time-varying and steady-state filters remains bounded by a constant. Similar to \cite{tsiamis9894660,rashidinejad2020slip,qian2025model}, we focus directly on the steady-state prediction in this paper.

\subsection{Optimal one-step prediction policy}

We briefly review here how to utilize the Kalman filter to build an autoregressive model for {\it single step} prediction \cite{tsiamis9894660}.  
Denote $e_k=y_k-\hat{y}_k$ as the innovation at time step $k$. By rolling out the Kalman filter \eqref{eq: KalmanPredictor} backwards for $p$ times, we~can reformulate  $y_{k+1}$ in terms of the past inputs and outputs~as
\begin{equation}\label{eq: oneStepAuto}
    y_{k+1}=\tilde{G}_{p} Z_{k, p}+C(A-L C)^{p} \hat{x}_{k-p+1}+e_{k+1},
\end{equation}
where
$
Z_{k, p} \triangleq\begin{bmatrix}
    y_{k-p+1}^{\tr}\!\!\!\!\!& \ldots & \!\!\!y_{k}^{\tr},u_{k-p+1}^\tr \ldots u_{k}^{\tr}
\end{bmatrix}^\tr
$ collects the past outputs and inputs, 
and $\tilde{G}_{p}$ denotes the optimal weights consisting of $\tilde{G}_p=\left[\tilde{G}_{1,p},\tilde{G}_{2,p}\right]$, where: 
\begin{subequations} 
\begin{align}
\label{eq:regressor}
\tilde{G}_{1,p} &\triangleq\begin{bmatrix}
C(A-L C)^{p-1} L, & \cdots, & C L
\end{bmatrix} \in \mathbb{R}^{m \times pm} \\
\label{eq:regressorInput}
 \tilde{G}_{2,p} &\triangleq\begin{bmatrix}
C(A-L C)^{p-1} B, & \cdots, & C B
\end{bmatrix} \in \mathbb{R}^{m \times pn_u}. 
\end{align}
\end{subequations}
This shows that the {\it optimal} steady-state policy for predicting one-step-ahead output is {\it a linear function of past outputs and inputs}. We here state another technical assumption {\!\!\cite{tsiamis9894660,rashidinejad2020slip,qian2025model}}: 
\begin{assumption} \label{asp:Diagonal}
    The matrix  $A-LC$ is diagonalizable.
\end{assumption}

\textcolor{black}{This assumption is used only to simplify the regret analysis. It ensures an exponential decay bound $\rho\left((A-LC)^p\right)\leq M\rho(A-LC)^p$ with some constant $M > 0$. There is no fundamental difficulty without \cref{asp:Diagonal}. If $A-LC$ contains higher-order Jordan blocks, the convergence speed of $(A-LC)^p$ will be $\rho\left((A-LC)^p\right)=\operatorname{poly}(p)\rho(A-LC)^p$. Hence, the backward horizon $p$ should be further extended to account for the slower convergence rate.}

\vspace{-1mm}
\subsection{Problem statement}
With observations up to time $k$, this paper aims to develop an online  \textit{multi-step-ahead} prediction policy of the form 
\begin{equation} \label{eq:multi-step-prediction}
\tilde{y}_{k+H}=f_{H}(y_0,\dots,y_{k},u_0,\dots,u_{k},\ldots,u_{k+H-1})
\end{equation}
where the term $\tilde{y}_{k+H}$ means the \textit{$H$-step ahead prediction} at time step $k$. If $H = 1$, this is reduced to the linear policy for \textit{one-step} prediction \eqref{eq: oneStepAuto}, as established in \cite{tsiamis9894660}.  
Following \cite{tsiamis9894660,rashidinejad2020slip,qian2025model}, we quantify the performance of the online prediction in terms of the \textit{regret} measured against the Kalman filter \eqref{eq: KalmanPredictor} that has full system knowledge. 
Under this setting, the original benchmark Kalman predictor from \eqref{eq: KalmanPredictor}, i.e., \[\hat{y}_{k+H}=\arg \min _{z \in \mathcal{F}_{k+H-1}} \mathbb{E}\left[\left\|y_{k+H}-z\right\|_{2}^{2} \mid \mathcal{F}_{k+H-1}\right],
\]  will be too strong, since it uses the information up to $y_{k+H-1}$. 

To address this, we consider a modified benchmark, called the \textit{$H$-step ahead Kalman predictor}, defined as
\begin{equation}\label{eq:MMSEProblemHstep}
    \bar{y}_{k+H} \triangleq \arg \min _{z \in \mathcal{F}_{k}} \mathbb{E}\left[\left\|y_{k+H}-z\right\|_{2}^{2} \mid \mathcal{F}_{k}\right].
\end{equation}
We aim to minimize the following regret
\begin{equation}\label{regret}
    \mathcal{R}_{N} \triangleq \sum_{k=1}^{N}\left\|y_{k+H}-\tilde{y}_{k+H}\right\|^{2}-\sum_{k=1}^{N}\left\|y_{k+H}-\bar{y}_{k+H}\right\|^{2},
\end{equation}
where $\tilde{y}_{k+H}$ is our online model-free prediction \eqref{eq:multi-step-prediction} and $\bar{y}_{k+H}$ is the optimal model-based Kalman prediction \eqref{eq:MMSEProblemHstep}.

We address two key questions in designing \eqref{eq:multi-step-prediction}: 1) how to parameterize the optimal multi-step-ahead prediction policy in terms of past outputs and control inputs; and 2) how to effectively learn this policy online and quantify its regret \eqref{regret} relative to the optimal model-based predictor.

%% file: Section-MainResult.tex
\section{Main results}
\subsection{Multi-step-ahead regression model}
We here derive a closed-form expression of the model-based optimal $H$-step ahead predictor and then provide a parameterization of the optimal prediction policy. 
\begin{lemma} \label{lemma:optimal-delayed-filter}
    Consider the linear stochastic system \eqref{eq: LinearSystem} and the optimal $H$-step ahead prediction problem \eqref{eq:MMSEProblemHstep}. The optimal $H$-step ahead prediction $\bar{y}_{k+H}$ can be obtained recursively as  \vspace{-1mm}
    \begin{equation}\label{eq: steadyPredictor}
    \begin{aligned}
       \bar{x}_{k+H}&=  A^{H-1} \hat{x}_{k+1}+  \sum_{i=1}^{H-1}A^{i-1}Bu_{k+i}, \\
       \bar{y}_{k+H}&=C\bar{x}_{k+H},
    \end{aligned}
\end{equation}
where $\hat{x}_{k+1}\triangleq \mathbb{E}\left\{x_{k+1}\mid \mathcal{F}_{k}\right\}$ is the standard Kalman state estimation from \eqref{eq: KalmanPredictor}. 
\end{lemma}

The proof is not difficult, and we present the details in \Cref{subsection:proof-multi-step-ahead-filter}. 
\Cref{lemma:optimal-delayed-filter} shows that the optimal $H$-step-ahead predictor can be obtained by first computing the best state estimate $\hat{x}_{k+1}$ via the standard Kalman filter and then propagating this estimate forward $H-1$ steps using the system dynamics together with the planned inputs $\{u_{k+1},\dots,u_{k+H-1}\}$. The predicted output $\bar{y}_{k+H}$ is simply the observation of this rolled-out state. Hence, optimal multi-step prediction reduces to applying the Kalman estimator once at the current time, followed by deterministic roll-out under the known dynamics. 

Lemma~\ref{lemma:optimal-delayed-filter} allows us to obtain an optimal prediction policy as a linear function of past outputs, past inputs, and future~inputs. 

\begin{theorem} \label{proposition:innovation}
    Let $r_{k+H}=y_{k+H}-\bar{y}_{k+H}$ denote the innovation process for the $H$-step ahead predictor \eqref{eq: steadyPredictor}. The following linear regression model holds 
\begin{equation}\label{eq: multiModel}
    y_{k+H}=G_pZ_{k,p}+CA^{H-1}(A-LC)^p\hat{x}_{k-p+1}+r_{k+H},
\end{equation}
where
$
Z_{k, p}\triangleq\begin{bmatrix}
    y_{k-p+1}^{\tr}\!\!\!\!\!& \ldots & \!\!\!y_{k}^{\tr},u_{k-p+1}^\tr \ldots u_{k+H-1}^{\tr}
\end{bmatrix}^\tr 
$ contains the past outputs, past inputs, and future inputs, 
and the regressor weights $G_{p}\triangleq\left[G_{1,p},G_{2,p}\right]$ are of the form: 
\[
\begin{aligned}
G_{1,p} \triangleq&\begin{bmatrix}
CA^{H-1}(A-L C)^{p-1} L, \!\!\!& \cdots, &\!\!\! CA^{H-1} L
\end{bmatrix},\\
    G_{2,p} \triangleq&\begin{bmatrix}
CA^{H-1}(A-L C)^{p-1} \!B, & \!\!\!\!\cdots\!, &\!\! CA^{H-1}B, \cdots\!, CB
\end{bmatrix}.
\end{aligned}
\]
Furthermore, we have 
\begin{equation} \label{eq:innovation}
r_{k+H}=e_{k+H}+\textstyle \sum_{i=1}^{H-1}CA^{i-1}Le_{k+H-i},     
\end{equation} 
where $e_k = y_{k} - \hat{y}_{k}$ is the innovation in the Kalman filter \eqref{eq: KalmanPredictor}.
\end{theorem}

We present the proof in \Cref{subsection:proof-linear-regression}. 
By definition, the $H$-step innovation $r_{k+H}$ aggregates past process noise $w_{k+1},\dots,w_{k+H-1}$. The overlap among these noise components induces correlations between $r_{k+H}$ and $r_{k+H-l}$ for $l< H$. Thus, the usual orthogonality of the one-step innovation process no longer holds in the $H$-step-ahead setting.  
As established in \eqref{eq:innovation}, this \textit{non-orthogonal} innovation $r_{k+H}$ can be parameterized as a combination of $e_{k+1}, \ldots, e_{k+H}$. 
   
In \cite{tsiamis9894660}, an $H$-step autoregressive (AR) model (without control inputs) is obtained by rolling out the one-step AR model $H$ times, and its innovation $\epsilon_{k+H}$ satisfies 
    $\epsilon_{k+H}=e_{k+H}+\sum_{i=1}^{H-1}CA^{i-1}Le_{k+H-i},
    $
    which coincides 
    with \eqref{eq:innovation}. 
    This observation highlights that including control inputs does not introduce additional uncertainty into the innovation process. Furthermore, \eqref{eq:innovation} provides an equivalent parameterization of $r_{k+H}$ as a summation of temporally uncorrelated one-step innovations $e_k$ across time. These properties are essential in our online learning algorithm and its regret analysis.  
 
\vspace{-1.5mm}
\subsection{Online learning and regret guarantee}
\vspace{-1mm}

From the linear regression \eqref{eq: multiModel}, the $H$-step-ahead output $y_{k+H}$ is a linear function of the past outputs, past inputs, and future inputs, perturbed by a bias term depending on $\hat{x}_{k-p+1}$ and an innovation process $r_{k+H}$. Following \cite{hazan2017learning, tsiamis9894660, rashidinejad2020slip, qian2025model}, we can estimate $G_p$ via least squares by regressing $y_{t+H}$ onto the regressor $Z_{t,p}, \;t \leq k$ (past inputs/outputs and future inputs).

\textbf{Least-squares}. We estimate ${G}_{k,p}$ by ridge regression: 
\begin{equation} \label{eq: leastsquare}
     \tilde{G}_{k,p}=\arg\min_{G } \,\,  \sum_{t=p}^{k-H} \left\|y_{t+H}-G{Z}_{t,p}\right\|_F^2+\lambda\left\| G\right\|_F^2, 
\end{equation}
where  $\lambda >0$  is a regularization parameter. At each time step $k$, by solving \eqref{eq: leastsquare}, we can obtain a closed form of $\tilde{G}_{k,p}$ as 
\begin{equation} \label{eq:regression-update}
   \tilde{G}_{k,p} = \textstyle \sum_{t = p}^{k-H} y_{t+H} Z_{t,p}^{\tr} V_{k-H,p}^{-1},  
\end{equation}
where  
  $  V_{k-
  H,p}\triangleq\lambda I + \sum_{t=p}^{k-H} Z_{t,p}Z_{t,p}^{\tr}$ 
is called the Gram~matrix, which contains all collected past available samples $Z_{t,p}$. 
 We then predict the future observation by  
\begin{equation} \label{eq:OPF-prediction}
    \tilde{y}_{k+H} = \tilde{G}_{k,p}Z_{{k},p}. 
\end{equation}

\textbf{Controlling bias}. For nonexplosive systems $\rho(A) \leq 1$, the bias term $\hat{x}_{k-p+1}$ in \eqref{eq: multiModel} retains the state from previous time steps, potentially growing at a polynomial rate. \textcolor{black}{Compared with the classical autoregressive model \cite{Tan2023Cooperative}, a persistent bias error $b_{k+H,p}\triangleq CA^{H-1}(A-LC)^p\hat{x}_{k-p+1}$ at each step $k$ could result in linear regret.} Fortunately, \Cref{lemma:optimal-delayed-filter} and classical theory guarantee that $\left\|A^{H-1}(A-LC)^{p}\right\|_2\leq c_1H^{\kappa-1}\rho(A-CL)^p$, with $\rho(A - LC) < 1$. This property allows us to control the accumulation of bias errors $A^{H-1}(A-LC)^{p}\hat{x}_{k-p+1}$ only with $p=\mathcal{O}(\log H\log k)$. 
We implement this via the standard doubling trick \cite{cesa2006prediction} (as in \cite{tsiamis9894660}): partition time into epochs of doubling length, and keep $p$ fixed within each epoch while increasing it between epochs. We note that the prediction policy in \cite{somalwar2025learning} chooses $p=1$, which will induce a significant bias and degrade the prediction performance.

\textbf{Recursive updates within an epoch}. At each time step $k$, we update the prediction $\tilde{y}_{k+H}$  using \eqref{eq:OPF-prediction} and then observe the new observation ${y}_{k+1}$. 
Within each epoch, the predictor can be computed recursively
\begin{subequations} \label{eq:predictor-update}
\begin{align}
V_{k-H,p}&=V_{k-H-1,p}+Z_{k-H, p} Z_{k-H, p}^{\tr},\\
\tilde{G}_{k,p}&=\tilde{G}_{k-1,p}+\left(y_{k}-\tilde{y}_{k}\right) Z_{k-H, p}^{\tr}V_{k-H,p}^{-1}.
\end{align}
\end{subequations}
A key distinction from the traditional one-step prediction~\cite{hazan2017learning, tsiamis9894660, rashidinejad2020slip, qian2025model} is the \textit{update delay}: 
the prediction \(\tilde y_{k+H}\) is not used immediately to update \(G\) at time \(k\); due to the \(H\)-step horizon, it contributes to the correction \(H\) steps later.

\vspace{-0pt}
\begin{algorithm}[t]
   \caption{H-step-ahead Online Prediction (\texttt{HOP})}\label{algPrediction}
\begin{algorithmic}[1]
   \STATE {\bfseries Input:} parameter $\beta, \lambda, T_{\text {init }}, N_E$ 
   
   \COMMENT{\textit{Warm Up}:}
   \FOR{$k=1$
   {\bfseries to }$T_{\text{init}}+1$}  
   \STATE Observe $y_{k}$, Generate $u_{k+H-1}$;
   \ENDFOR
   
   \COMMENT{\textit{Recursive Online Prediction:}}
    \FOR{$l=1$ {\bfseries to} $N_E$}
    \STATE
    Initialize 
      $T_l\!=\!2^{l-1} T_{\text {init }}+1, p\!=\!\beta \log T_l,$ 
    \STATE
    Compute $V_{T_l-H,p}$ and  $\tilde{G}_{T_l,p}$; 
    \FOR{$k=T_l$ \textbf{to} $2 T_l-2$}
    \STATE Predict $\tilde{y}_{k+H}=\tilde{G}_{k,p} Z_{k,p}$; 
    \STATE
    Observe $y_{k+1}$, Generate $u_{k+H}$;
    \STATE
    Update $V_{k-H+1,p}$ and $\tilde{G}_{k+1,p}$ as \eqref{eq:predictor-update}. 
    \ENDFOR
    \ENDFOR   
\end{algorithmic}
\end{algorithm}

The model-free multi-step-ahead prediction is summarized in \Cref{algPrediction}.  We have the following regret guarantee. 
\begin{theorem}\label{thm1}
    Consider the linear stochastic system \eqref{eq: LinearSystem}. Suppose \Cref{asp: System,asp:Diagonal} hold. For a fixed  $H$-step-ahead prediction, we choose the parameters in \Cref{algPrediction}  as  
    \begin{equation}\label{eq: beta}
        \beta=\frac{\mathcal{O}(\kappa+\log H)}{\log (1 / \rho(A-L C))},\quad
    \end{equation}
    where $\kappa$ represents the order of the largest Jordan block of eigenvalue 
$1$ in matrix $A$. 
 Almost for sure, we have 
    \begin{equation} \label{eq:final-regret-bound}
    \mathcal{R}_{N} \leq MH^{4\kappa+1}\beta^3 \mathcal{O}\left(\log^{7} N\right),
    \end{equation}
   where $M$ is a constant related to the system parameters.
\end{theorem}

We outline the proof in \Cref{sec: proofSketch} and highlight the key technical differences compared with the literature \cite{hazan2017learning, tsiamis9894660, rashidinejad2020slip, qian2025model}.

 \cref{thm1} provides the \textit{first} logarithmic regret for online multi-step-ahead prediction with respect to the optimal model-based predictor.
 When $A$ has Jordan blocks at eigenvalue $1$, the regret remains logarithmic in 
$N$, but its constant scales polynomially with the prediction horizon $H$, with degree determined by the size of the largest such block. This logarithmic regret holds true despite the innovation $r_{k+H}$ being non-orthogonal.  Compared with \cite{hazan2017learning, tsiamis9894660, rashidinejad2020slip, zhang2023learningTK, qian2025model}, our result gives the first explicit polynomial scaling of the regret in the prediction horizon~$H$. 

Moreover, our bound holds almost surely, i.e., it does not depend on a fixed failure probability $\delta \in (0,1)$. 
\textcolor{black}{The key idea for this property is that for fixed $\delta$, the stochastic confidence term of the original regret in \cite{tsiamis9894660,rashidinejad2020slip,qian2025model} will scale with $\poly\log \frac{1}{\delta}$.  For sufficiently large $N$, we only need to let $\delta=\frac{1}{N}$, then the stochastic confidence term can be dominated by the information accumulation term, i.e., $\log\det V_{k,p}$. This property coincides with the empirical result that the norm of a long Gaussian random process is ``very close" to the norm of its expectation \cite{vershynin2018high}, and it also provides new insights into how a large amount of data can suppress the impact of randomness in the learning process.} To the best of our knowledge, this is the first almost sure bound for online prediction, although a similar technique has also been utilized in online LQR \cite{lu10857470}.

\section{Technical Proofs}\label{sec: proofSketch}

\subsection{Proof of \Cref{proposition:innovation}} \label{subsection:proof-linear-regression}
      {First, since $\hat{x}_{k+1}\triangleq \mathbb{E}\left\{x_{k+1}\mid \mathcal{F}_{k}\right\}$ is the standard Kalman's state estimation from \eqref{eq: KalmanPredictor}, we can roll the standard Kalman filter backwards for $p$ times and get $\hat{x}_{k+1}=(A-LC)^p\hat{x}_{k+1-p}+\sum_{l=0}^{p-1}(A-LC)^{l}(Ly_{k-l}+Bu_{k-l})$. Then the regression model \eqref{eq: multiModel} is a direct consequence of the recursive update \eqref{eq: steadyPredictor} combined with the above recursive relation.}

    In the following, we establish the relationship \eqref{eq:innovation}, and only consider the steady-state innovation process. 
    Denote $\tilde{x}_{k+1}\triangleq x_{k+1}-\hat{x}_{k+1}$ as the one-step optimal state prediction error for Kalman filter \eqref{eq: KalmanPredictor}, then we have 
    $\mathbb{E}\left\{\tilde{x}_{k+1}\tilde{x}_{k+1}^\tr\right\}=P,\quad\forall k\in\mathbb{N},$ where $P$ is from \eqref{eq: DARE}, and the recursion
    $
    \tilde{x}_k=(A-LC)\tilde{x}_{k-1}+w_{k-1}-Lv_{k-1} 
    $ holds.
    
    By comparing $y_{k+H}$ and $\bar{y}_{k+H}$ directly, we obtain
    \begin{equation}\label{eq: essentialR}
        r_{k+H}=CA^{H-1}\tilde{x}_{k+1}+v_{k+H}+\sum_{i=1}^{H-1}CA^{i-1}w_{k+H-i}.
    \end{equation}
    Consider the structure of $e_k=C\tilde{x}_k+v_k$. We have
    $
        e_{k+H}\!=\!C(A\!-\!LC)\tilde{x}_{k+H-1}\!\!+\!Cw_{k+H-1}\!-\!CLv_{k+H-1}\!+\!v_{k+H}.
   $
    Together with 
    $
    CLe_{k+H-1}=CLC\tilde{x}_{k+H-1}+CLv_{k+H-1},
    $ 
    we further have
    $e_{k+H}+CLe_{k+H-1}=CA\tilde{x}_{k+H-1}+v_{k+H}+Cw_{k+H-1}.$
    With a similar procedure, we have a recursive deduction below  
    \[
    \begin{aligned}
    &e_{k+H}+ \textstyle\sum_{i=1}^{H-1}CA^{i-1}Le_{k+H-i}
    \\
        =&CA\tilde{x}_{k+H-1}+v_{k+H}+Cw_{k+H-1}+\textstyle\sum_{i=2}^{H-1}CA^{i-1}Le_{k+H-i}\\
        =&CA^{H-1}\tilde{x}_{k+1}+v_{k+H}+\textstyle\sum_{i=1}^{H-1}CA^{i-1}w_{k+H-i}\end{aligned}.
    \]
    Combining this with \eqref{eq: essentialR} leads to the desired relationship \eqref{eq:innovation}. 

\vspace{-2pt}
\subsection{Proof of \Cref{thm1}}
We outline the proof of \Cref{thm1} in four main steps. More details for each step are provided in \Cref{appendix:decomposition,appendix:bias-factor,sec:proofRegression,appendix:accumulation}. 

\subsubsection{Decomposition of regret} With \cite[Theorem 1]{tsiamis9894660}, the regret $\mathcal{R}_N$ is dominated by $\mathcal{L}_N\triangleq \sum_{k=T_{\textnormal{init}}}^{N}\left\|\tilde{y}_{k+H}-\bar{y}_{k+H}\right\|_{2}^{2}$, i.e., $\mathcal{R}_N=\mathcal{L}_N+o(\mathcal{L}_N)$. Hence, in the following proof, we mainly analyze the scaling law of $\mathcal{L}_N$ with respect to $N$ and $H$. We further decompose each $\tilde{y}_{k+H}-\bar{y}_{k+H}$ into three parts,
    \[
    \begin{aligned}       
   & \tilde{y}_{k+H}-\;\hat{y}_{k+H}=  \underbrace{\sum_{l=p}^{k-H} b_{l+H, p} Z_{l, p}^{\tr} V_{k-H, p}^{-1} Z_{k, p}-b_{k+H, p}}_{\text{Bias error}}  
   \\&\quad+\underbrace{\sum_{l=p}^{k-H} r_{l+H}Z_{l, p}^{\tr} V_{k-H, p}^{-1} Z_{k, p}}_{\text{Regression error}}-\underbrace{\lambda G_{p} V_{k-H, p}^{-1} Z_{k, p}}_{\text{Regularization error}}. 
    \end{aligned}
    \]

\vspace{-2mm}
    
    We further denote $\mathcal{B}_{k,p}\triangleq \Big\| \sum_{l=p}^{k-H} b_{l+H, p} Z_{l, p}^{\tr} V_{k-H, p}^{-\frac{1}{2}}\Big\|_{2}^{2}$ as the bias factor, $\mathcal{E}_{k,p}\triangleq \Big\| \sum_{l=p}^{k-H} r_{l+H, p} Z_{l, p}^{\tr} V_{k-H, p}^{-\frac{1}{2}}\Big\|_{2}^{2}$ as the regression factor, $\mathcal{G}_{k,p}\triangleq \Big\| \lambda G_{p} V_{k-H, p}^{-\frac{1}{2}}\Big\|_{2}^{2}$ as the regularization factor, $\mathcal{V}_{N,p}\triangleq\!\!
    \sum_{k=T_{\textnormal{init}}}^{N}\left\|V_{k-H, p}^{-\frac{1}{2}} Z_{k,p}\right\|_{2}^{2}$
    as the accumulation factor, and $
\mathfrak{b}_{N+H,p}=\sum_{k=T_{\textnormal{init}}}^{N}\left\|b_{k+H, p}\right\|_{2}^{2}$ as the accumulation of bias. Then we can further decompose the term $\mathcal{L}_N$ into
\begin{equation}\label{eq: decoupleRegretBound}
        \mathcal{L}_{N}\leq 4\big(\max_{k\leq N
}\left(\mathcal{B}_{k,p} +\mathcal{E}_{k,p} + \mathcal{G}_{k,p}\right)\big)\cdot\mathcal{V}_{N,p}+4\mathfrak{b}_{N+H,p}.
 \end{equation}
    Note that from the expression of $G_p$, we can obtain that 
   $
    \mathcal{G}_{k,p}\leq \frac{MH^{2\kappa}}{1-\rho(A-LC)}
    $ does not scale with $N$. 
    Thus, in the following proof, we discuss the bias factor $\mathcal{B}_{k,p}$, the regression factor $\mathcal{E}_{k,p}$, and the accumulation factor $\mathcal{V}_{N,p}$.

\subsubsection{Term-wise Analysis of the Regret} 
  In this part, we provide a detailed analysis of each term in the regret bound \cref{eq: decoupleRegretBound}. The following lemmas establish the term-wise bounds.
    \begin{lemma} \label{lemma:bias-factor}
Suppose \Cref{asp: System,asp:Diagonal} hold. For any fixed $\beta$ satisfying the condition in \Cref{thm1}, we have 
    \begin{equation}  \label{eq:bias-factor}
\mathfrak{b}_{N+H,p}\!\leq\! M\!\log^2N,\;\;\text{and}\;\;\max_{T_{\textnormal{init}}\!\leq k\! \leq N}\mathcal{B}_{k,p}\!\leq \!M\!\log N
\end{equation}
holds almost for sure.
\end{lemma}

\begin{lemma} \label{lemma:regression-factor}
Suppose \Cref{asp: System} holds. For any fixed $\beta$ satisfying the condition in \Cref{thm1}, we have 
    \begin{equation} 
\max_{T_{\textnormal{init}}\leq k \leq N}\mathcal{E}_{k,p}\!\leq \!MH^{2\kappa}\beta\log^2 N
\end{equation}
holds almost for sure.
\end{lemma}

\begin{lemma} \label{lemma:accumulation-factor}
Suppose \Cref{asp: System} holds. For any fixed $\beta$ satisfying the condition in \Cref{thm1}, we have 
    \begin{equation} 
\mathcal{V}_{N,p}\!\leq \!MH^{2\kappa+1}\beta^2\log^5 N
\end{equation}
holds almost for sure.
\end{lemma}

\textcolor{black}{The proof ideas of \cref{lemma:bias-factor,lemma:regression-factor} generally follow the procedure in \cite[Theorem 1]{tsiamis9894660} and \cite[Theorem 1]{abbasi2011improved}, with additional derivations to handle the accumulation effect induced by multi-step-ahead prediction. The detailed proofs are provided in the \Cref{appendix:bias-factor,sec:proofRegression,appendix:accumulation}. One of the key distinctions of our analysis lies in the treatment of the stochastic confidence term, which requires careful examination of the dominance relationship between stochastic confidence and information accumulation within each component of the regret bound.}
\textcolor{black}{The proof of \cref{lemma:accumulation-factor} contains several non-trivial mathematical techniques. To highlight the core theoretical contribution, we present the essential proof steps of \cref{lemma:accumulation-factor} in the next subsection.}  

\textbf{Proof of \Cref{thm1}}: We can now directly combine \eqref{eq: decoupleRegretBound} and \Cref{lemma:bias-factor,lemma:regression-factor,lemma:accumulation-factor} to establish the desired regret bound in \eqref{eq:final-regret-bound}, where we have eliminated the low-order terms.

\subsection{Proof of \cref{lemma:accumulation-factor}}
Without loss of generality, we directly assume $N=2^{N_E} T_{\text{init}}$, where $N_E$ is the number of epochs. Then
    we can decompose the $\mathcal{V}_{N,p}$ into 
    $
\mathcal{V}_{N,p}\!\leq\!\! \max_{T_{\text{init}}\leq k\leq N}\left\|V_{k-H,p}^{-\frac{1}{2}}V_{k,p}^{\frac{1}{2}}\right\|_2^2\times\sum_{l=1}^{N_E}\!\sum_{k=T_l}^{2T_l-2}\!\left\|V_{k, p_l}^{-\frac{1}{2}} Z_{k,p_l}\right\|_{2}^{2},
$
where the subscript of $p_l$ is to highlight the $p$ varies with epoch number $l$.
The inequality
$
\textstyle \sum_{k=T_{l}}^{2T_l-2}\left\|V_{k, p_l}^{-\frac{1}{2}} Z_{k,p_l}\right\|_{2}^{2}\leq \log\big(\det V_{2T_l-2,p_l}/\det V_{T_l-1,p_l}\big)
$ 
 directly holds with \cite[Lemma 1]{tsiamis9894660}.

\textcolor{black}{The nontrivial part lies in the analysis for the dependence of $\max_{T_{\text{init}}\leq k\leq N}\big\|V_{k-H,p}^{-\frac{1}{2}}V_{k,p}^{\frac{1}{2}}\big\|_2^2$ on $H$. Note that we have 
$
\big\|V_{k-H,p}^{-\frac{1}{2}}V_{k,p}^{\frac{1}{2}}\big\|_2^2
    \leq 1 + \sum_{l=k-H+1}^k Z_{l,p}V_{k-H,p}^{-1}Z_{l,p}^\tr,
$
which cannot be directly bounded with system parameters. 
Based on the innovation representation of output $y_k$ as
$$
y_k=CA^{d}\hat{x}_{k-d}+\sum_{i=1}^{d}CA^{i-1}Bu_{k-i}\!+\!e_k\!+\!\sum_{i=1}^{d}CA^{i-1}Le_{k-i}.
$$
We can derive the $H$-step AR representation of $Z_{k,p}$ as
\[
Z_{k,p}=\sum_{i=1}^{d}a_{i-1}^{(H)} Z_{k-H+i-d,p}+\delta_{k,p}+\sum_{j=1}^{H-1}a_{d-1}^{(j)}\delta_{k-j,p},
\] 
where each $a_{i-1}^{(j)}$ is calculated with rolling out minimal polynomial of $A$ for $j$ times, i.e., $A^d = a_{d-1}^{(j)}A^{d-j}+\dots+a_0^{(j)}$, and the degree of the polynomial is $d$. With the fact $Z_{l,p}^\tr V_{s,p}^{-1}Z_{l,p}\leq 1,\; \forall l\leq s$, we can further decompose the quadratic form $Z_{l,p}V_{k-H,p}^{-1}Z_{l,p}^\tr,\;k-H+1\leq l\leq k$ as 
\begin{align}
     &Z_{l,p}^\tr V_{k-H,p}^{-1}Z_{l,p}\!\leq\! d(H\!+\!1)\!\sum_{i=1}^{d}\big(a_{i-1}^{(H)}\big)^2\!\!+\!(H\!+\!1)\delta_{l,p}^\tr V_{k-H,p}^{-1}\delta_{l,p}\nonumber\\
&\qquad\quad+(H+1)\sum_{j=1}^{H-1}a_{d-1}^{(j)}\delta_{l-j,p}^\tr V_{k-H,p}^{-1} a_{d-1}^{(j)}\delta_{l-j,p}.
\end{align}
Each term $\delta_{k,p}$ contains the information of decoupled innovation $e_s$ and control input $u_s$ with $s\leq k$, we can further show that $\max_{k\leq N}\left\|\delta_{k,p}\right\|_{2}^2\leq M\beta \log^2 N$ holds almost for sure.
Then the key challenge here is to determine how $a_{i-1}^{(H)}$ scales with $H$, as compound effects could potentially cause exponential growth. 
Fortunately, $a_{i-1}^{(j)}$ arises from the last column of the matrix $\mathcal{A}^j$, where $\mathcal{A} =
\left[\begin{smallmatrix}
0  & 0 & \cdots & 0 & a_0 \\
1 & 0 & \cdots & 0 & a_1 \\
\vdots & \vdots & \vdots & \ddots & \vdots \\
0 & 0 & \cdots & 1 & a_{d-1}
\end{smallmatrix}\right]$ is the companion matrix of $A$.
Thus, $\mathcal{A}$ shares the same minimal polynomial as \(A\), and we have  $\|\mathcal{A}^j\|_F^2 \le M j^{2\kappa-2}$, implying $(a_{i-1}^{(j)})^2 \le M j^{2\kappa-2}$ uniformly for all $i \le d$ and $j \le H$. This confirms that the coefficients scale {polynomially} with the prediction horizon $H$.  We present further details in \Cref{appendix:accumulation}.} 

%% file: Section-Simulation.tex
\section{Numerical experiments}
We here provide numerical experiments to verify the performance of the proposed \texttt{HOP} in \Cref{algPrediction}.
We consider a modified dynamical system model from \cite[Section V]{tsiamis9894660} with control inputs. The system parameters are given by 
$$
A=\begin{bmatrix}
1 & 0.5 & 0 \\
0 & 1 & 0.5 \\
0 & 0 & 0.9
\end{bmatrix},\;\; B=\begin{bmatrix}
    0\\0\\1
\end{bmatrix},\;\; C=\begin{bmatrix}
    1&0&0
\end{bmatrix},
$$ 
and $Q=0.01*I_3, R=0.01$. The control input $u_k$ is randomly generated from i.i.d. standard Gaussian distribution, i.e., $u_k\sim \mathcal{N}(0,1)$. The hyperparameter is chosen to be $\beta=2$, and $T_{\text{init}}=400$, the number of epochs is 3.

\begin{figure}[t]
    \centering
    \setlength{\abovecaptionskip}{2pt}
    \includegraphics[height=0.3\textwidth]{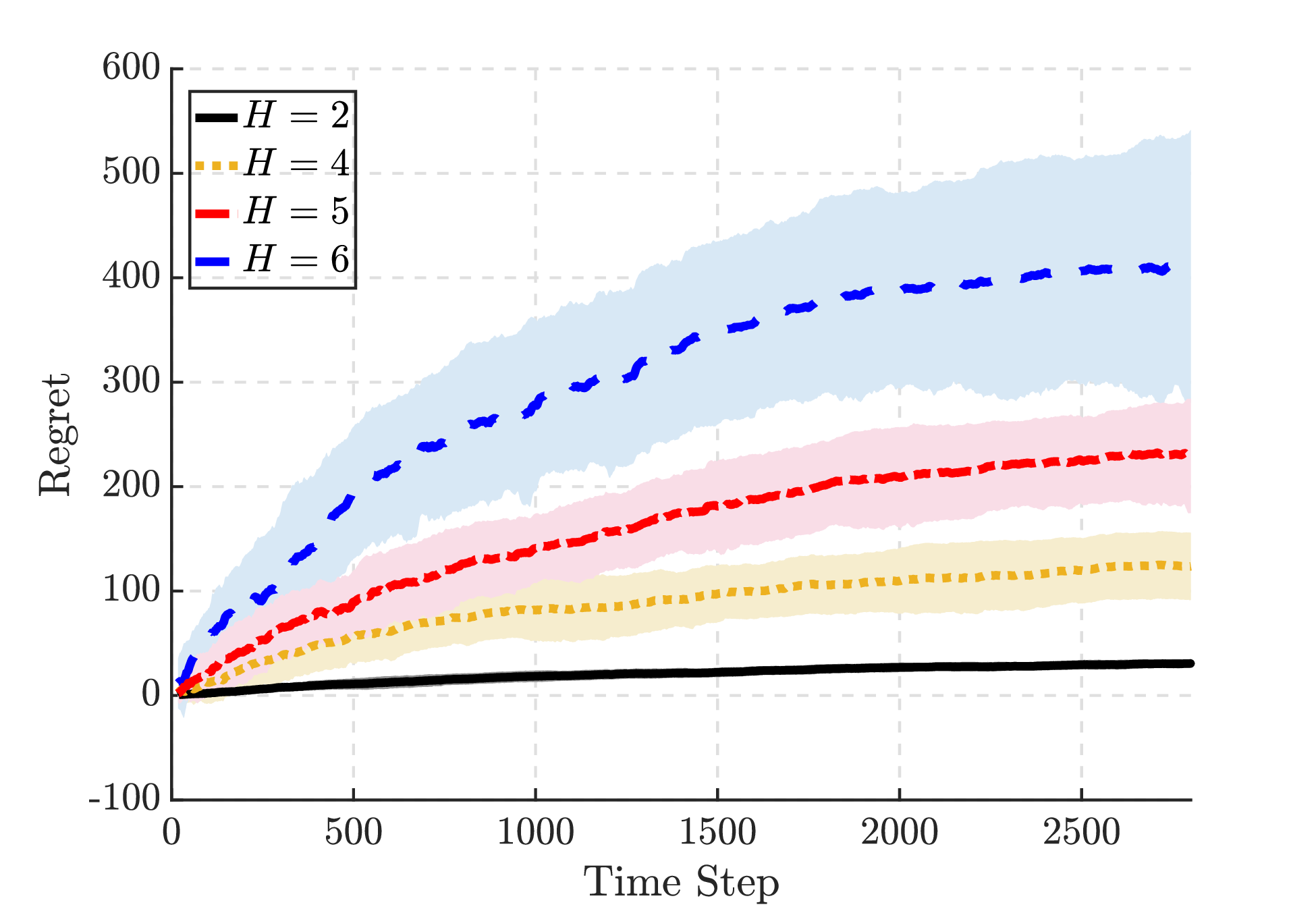}
    \caption{Comparison of regret $\mathcal{R}_N$ across prediction horizons~$H\in \{2,4,5,6\}$. The regret remains logarithmic for all $H$. The multiplicative constant increases nonlinearly with $H$.} 
    
    \vspace{-3mm}
    \label{figComparison}
\end{figure}

In \Cref{figComparison}, we provide the comparison of regret $\mathcal{R}_N$ with different ahead prediction step $H$, where $H$ are chosen to be $2,4,5,6$ respectively. We can see that the regret remains logarithmic despite different $H$. For a marginally stable system, the regret scales nonlinearly with the increase of $H$, which is consistent with the polynomial scaling claim in \cref{thm1}.

To further verify the scaling property of $\mathcal{R}_N$ with respect to $H$, we also consider an open-loop stable system. 
The system matrix $A$ is chosen to be 
$
A=\left[\begin{smallmatrix}
0.6 & 0.5 & 0 \\
0 & 0.6 & 0.5 \\
0 & 0 & 0.6
\end{smallmatrix}\right],
$
and the parameters $B,C,Q,R$ remain the same as those in the previous experiment. The regrets for the marginally stable system and the open-loop stable system with different $H$ are listed in \Cref{tab:comparison}.
\renewcommand{\arraystretch}{1.3}
\begin{table}[t]
\centering
\caption{Comparison of $R_N$ scaling with $H$ for marginally stable systems and open-loop stable systems.}
\begin{tabular}{cccccccc}
\toprule
{Ahead step $H$} & 2 & 4 &6&8&10&12 \\
\hline
$\mathcal{R}_N$ ($\rho(A)=1$) & 30.7 & 123.7 & 410.9 & 1035&2280&4600\\
$\mathcal{R}_N$ ($\rho(A)=0.6$)
& 2.84 & 3.49 & 3.60 & 4.48 & 5.08 & 4.78 \\
\bottomrule
\end{tabular}
\label{tab:comparison}
\end{table}
We can see that for the open-loop stable system, the regret $\mathcal{R}_N$ scales roughly linearly with $H$, and even experiences a saturation effect for large enough $H$. While for marginally stable systems, $\mathcal{R}_N$ scales polynomially with $H$, with the polynomial order between $H^2$ and $H^3$. These results are consistent with our analysis in \cref{thm1}, though the polynomial exponent there may be conservative.

%% file: Section-Proof.tex
\section{Additional details}\label{sec:technical}

Before we introduce the detailed proof for each of the Lemmas, we provide a brief summary of the main challenge for the proof and the key technical advance. 
For the theoretical analysis part, the proof of the \textit{polynomial scaling} of the regret with respect to $H$ requires a new understanding of the \textit{multi-step autoregressive representation} of $Z_{k,p}$ in terms of past samples $Z_{k-H-i,p},\; i \ge 0$. The key challenge is to show the parameters $a_{i}^{(H)}, 0\leq i\leq d-1$ in the autoregressive model scale polynomially with $H$. Moreover,
the \textit{almost-sure bound} for the regret that is independent of the probability parameter $\delta$, also requires analyzing the dominance between the \textit{stochastic confidence} term $\poly \log\frac{1}{\delta}$ and \textit{information accumulation} term $\log\det V_{k,p}$ in each part of the regret bound. 

\subsection{Proof of \Cref{lemma:optimal-delayed-filter}} \label{subsection:proof-multi-step-ahead-filter}
    It is known that the optimal prediction \eqref{eq:MMSEProblemHstep} is equivalent to the conditional expectation \cite[Section 2]{Andersonoptimal}, i.e., 
    $$\bar{y}_{k+H} = \mathbb{E}\left\{y_{k+H}\mid \mathcal{F}_{k}\right\}.$$ 
    We thus only need to compute this condition expectation.~
Let $f(x)$ be the probability distribution function of a random variable $x$. 
The Chapman-Kolmogorov (C-K) equation says that 
     \begin{equation}\label{eq: propagation}
         f\left(x_{k+2} \!\mid\!\! \mathcal{F}_{k}\right)=\int\! f\left(x_{k+2}\! \mid\! x_{k+1}\right) f\left(x_{k+1} \!\mid\! \mathcal{F}_{k}\right) \mathrm{d} x_{k+1}.
     \end{equation}
    From the system dynamics \eqref{eq: LinearSystem}, we have
    $$f\left(x_{k+1} \!\mid\! x_{k}\right)=\mathcal{N}\left(Ax_{k}+Bu_{k},Q\right).$$ 
    Classical Kalman filtering theory guarantees that the steady-state estimation \cite{Andersonoptimal} satisfies 
    $$
    f\left(x_{k+1} \mid \mathcal{F}_{k}\right)=\mathcal{N}\left(\hat{x}_{k+1},P\right),
    $$
    where $P$ is the unique positive semidefinite solution to the ARE \eqref{eq: DARE}.
    Calculating the integral \eqref{eq: propagation}, we have
    \[
    f\left(x_{k+2} \mid \mathcal{F}_{k}\right)=\mathcal{N}\left(A\hat{x}_{k+1}+Bu_{k+1},APA^\tr+Q\right).
    \]
    Performing the integral \eqref{eq: propagation} for $H-1$ times recursively leads to 
    $$
    f\left(x_{k+H} \mid \mathcal{F}_{k}\right)=\mathcal{N}\left(\bar{x}_{k+H},P_H\right),
    $$
    where the mean $\bar{x}_{k+H}$ is defined in \eqref{eq: steadyPredictor} and the covariance $P^{(H)}$ takes the form as
    \begin{equation}\label{eq: multivariance}
        P_H=A^{H-1} P\left(A^{H-1}\right)^\tr+\textstyle \sum_{i=1}^{H-1}A^{i-1} Q (A^{i-1})^\tr.
    \end{equation}
    Since we have $\mathbb{E}\left\{y_{k+H}\mid \mathcal{F}_{k}\right\}=C\mathbb{E}\left\{x_{k+H}\mid \mathcal{F}_{k}\right\}$, the proof is now completed.

\subsection{Details for Regret decomposition}\label{appendix:decomposition}
 
 With classical results in online linear regression techniques \cite{tsiamis9894660, rashidinejad2020slip}, we first divide the regret $\mathcal{R}_N$ into two parts, i.e.,\vspace{-5pt}
\[
\begin{aligned}
    \mathcal{R}_{N} \triangleq& \sum_{k=T_{\text {init }}}^{N}\left\|y_{k+H}-\tilde{y}_{k+H}\right\|_2^{2}\;-\sum_{k=T_{\text {init }}}^{N}\left\|y_{k+H}-\bar{y}_{k+H}\right\|_2^{2} \nonumber \\
    =&\underbrace{\sum_{k=T_{\text {init }}}^{N}\!\!\!\left\|\bar{y}_{k+H}-\tilde{y}_{k+H}\right\|_{2}^{2}}_{\mathcal{L}_{N}}\!+2 \!\underbrace{\sum_{k=T_{\text {init }}}^{N}\! r_{k+H}^{\tr}\!\left(\bar{y}_{k+H}\!-\!\tilde{y}_{k+H}\right)}_{\text {cross term }} .
\end{aligned}\]
The first part is the accumulation of the gap $\bar{y}_{k}-\tilde{y}_{k}$, while the second part is a cross term between the innovation $r_{k+H}$ and the gap $\bar{y}_{k}-\tilde{y}_{k}$. Since $r_{k+H}$ can be decoupled into the summation of $H$ i.i.d. Gaussian sequences, from the self-normalized martingale theory \cite{abbasi2011improved}, it is standard~to~bound
\begin{equation} \label{eq:bound-marignale-term}
\sum_{k=T_{\text {init }}}^{N} r_{k+H}^{\tr}\left(\bar{y}_{k+H}-\tilde{y}_{k+H}\right)=\tilde{O}\left(\sqrt{\mathcal{L}_N}\right)=o\left(\mathcal{L}_N\right),
\end{equation}
i.e., the cross term is dominated by the accumulation term $\mathcal{L}_N$.

Then following standard linear regression techniques \cite{abbasi2011improved, hazan2017learning}, we can divide the gap $\bar{y}_{k+H}-\tilde{y}_{k+H}$ at each time step $k$ as:  
\begin{itemize}
    \item  1) the \textit{regularization} error induced by $\lambda I$, 
\item 2) the \textit{regression} error induced by Gaussian innovation $r_{k+H}$, 
\item 3) the \textit{bias} error induced by $b_{k+H,p}=CA^{H-1}(A-K C)^{p} \hat{x}_{k-p+1}$. 
\end{itemize}
In particular, we rewrite the gap $\tilde{y}_{k+H}-\bar{y}_{k+H}$ as 
\begin{align}
   & \tilde{y}_{k+H}-\;\hat{y}_{k+H}=  \underbrace{\sum_{l=p}^{k-H} b_{l+H, p} Z_{l, p}^{\tr} V_{k-H, p}^{-1} Z_{k, p}-b_{k+H, p}}_{\text{Bias error}} \nonumber\\&\quad+\underbrace{\sum_{l=p}^{k-H} r_{l+H}Z_{l, p}^{\tr} V_{k-H, p}^{-1} Z_{k, p}}_{\text{Regression error}}-\underbrace{\lambda G_{p} V_{k-H, p}^{-1} Z_{k, p}}_{\text{Regularization error}}. \label{eq:gap-yk-decomposition}
\end{align}   
The main difference between  \eqref{eq:gap-yk-decomposition} and the decomposition in \cite{tsiamis9894660, rashidinejad2020slip} is that the innovation $r_{k+H}$ is inherently correlated, i.e., $\mathbb{E}\left\{r_{k+H}r_{k+H-l}^{\tr}\right\}\neq 0, \forall l<H$, which requires some decoupling techniques. Due to the time delay, the bias term will be affected by $A^{H-1}$, which requires a longer past horizon to suppress. Furthermore, the delay induced asymmetry, i.e., the cross between $V_{k-H, p}^{-\frac{1}{2}} Z_{k, p}$ will further complicate the analysis process.
Following a simple argument, we have the following bound. 
\begin{align} \label{wholeRegret}
\mathcal{L}_{N}=&\sum_{k=T_{\textnormal{init}}}^{N}\left\|\tilde{y}_{k+H}-\bar{y}_{k+H}\right\|_{2}^{2} \nonumber\\
\leq&4\big(\max_{k\leq N
}\left(\mathcal{B}_{k,p} +\mathcal{E}_{k,p} + \mathcal{G}_{k,p}\right)\big)\cdot\mathcal{V}_{N,p}+4\mathfrak{b}_{N,p}, 
\end{align}
where the factors are defined as 
\[
\begin{aligned}
    \mathcal{B}_{k,p}\!\triangleq& \Big\| B_{k,p}\bar{Z}_{k-H,p} \!V_{k-H, p}^{-\frac{1}{2}}\Big\|_{2}^{2}, \mathcal{E}_{k,p}\!\triangleq \! \left\|\mathcal{R}_{k, p} \bar{Z}_{k-H, p}^{\tr} \!V_{k-H, p}^{-\frac{1}{2}}\right\|_{2}^{2}\!,\\
   \mathcal{G}_{k,p}\triangleq&\;\Big\| \lambda G_{p} D_{p}^{-2} V_{k-H, p}^{-\frac{1}{2}}\Big\|_{2}^{2}, \;\; \mathcal{V}_{N,p}\triangleq\!\!
\sum_{k=T_{\textnormal{init}}}^{N}\left\|V_{k-H, p}^{-\frac{1}{2}} Z_{k,p}\right\|_{2}^{2},
\end{aligned}
\]  
 $
\mathfrak{b}_{N+H,p}=\sum_{k=T_{\textnormal{init}}}^{N}\left\|b_{k+H, p}\right\|_{2}^{2}$.
In the terms above, $B_{k,p}\triangleq\begin{bmatrix}
    b_{p+H,p},\dots,b_{k,p}
\end{bmatrix}$, $\mathcal{R}_{k,p}\triangleq\begin{bmatrix}
    r_{p+H},\dots,r_k
\end{bmatrix},$ $\bar{Z}_{k-H,p}\triangleq\begin{bmatrix}
    Z_{p,p},\dots,Z_{k-H,p}
\end{bmatrix}$ are the collections of all past bias $b_{l,p}$, innovations $r_l$, and samples $Z_{l,p}$, respectively \footnote{For a specific $k$ in different epochs, the parameter $p$ will be different due to the doubling trick. In \Cref{algPrediction}, we always have $p\leq \beta\log k$.}. Without loss of generality, we directly assume $N=2^{N_E} T_{\text{init}}$, where $N_E$ is the number of epochs.

\subsection{Proof for \cref{lemma:bias-factor} and selection of $\beta$} \label{appendix:bias-factor}

We first provide an almost sure bound for the bias factors and derive the requirement for $\beta$. 
Note that at each epoch $l$, for any $T_l\leq k\leq 2T_l-2$ there is 
$$
\begin{aligned}
    \mathcal{B}_{k,p}=& \left\|B_{k, p} \bar{Z}_{k-H, p}^{{\tr}} \bar{V}_{k-H, p}^{-1}\bar{Z}_{k-H, p} B_{k,p}^{\tr}\right\|_2^2
    \leq \sum_{i=p+H}^{k}\left\|b_{i, p}\right\|_{2}^{2}\\\leq&\sum_{i=p+H}^{k}\left\|CA^{H-1}(A-LC)^p\right\|_2^2\left\|\hat{x}_{i-p-
    H+1}\right\|_2^2,
\end{aligned}
$$
where $p=\beta \log(T_l-1)$.
 Denote $\Gamma_k\triangleq \mathbb{E}\left\{\hat{x}_k\hat{x}_k^\tr\right\}$, with \Cref{lm: Deviation}, we have that for a fixed probability $\delta$ and for all $k\in\mathbb{N}$, there is
\[
\begin{aligned}
    \left\|\hat{x}_{k}\right\|_2^2\leq& \left(2n+3\log \frac{k^2}{\delta}\right)\left\|\Gamma_k\right\|_2^2\\
    \leq& \left(2n+3\log \frac{k^2}{\delta}\right)\big(\left\|Q\right\|_2+\left\|B\right\|_2^2\big)\sum_{i=0}^{k-1}\left\|A^i\right\|_2^2
\end{aligned}
\]
uniformly holds for all $k$ with probability at least $1-\frac{\pi^2\delta}{6}$.
Note that $\left\|A^i\right\|_2^2\leq  Mi^{2\kappa-2}$, where $M$ is a constant only related to system parameters, together with $\left\|CA^{H-1}(A-LC)^p\right\|_2^2\leq MH^{2\kappa-2}\rho(A-LC)^{2p}$, we first have that the inequality
\[
\mathcal{B}_{k,p}\leq M H^{2\kappa-2}\rho(A-LC)^{2p} k^{2\kappa} \log\frac{1}{\delta}, 
\]
holds for all $k$ uniformly with probability $1-\frac{\pi^2\delta}{6}$.
The above inequality holds due to $\log k\leq k$.
We then choose $\beta=\frac{M_1}{\log(1/\rho(A-LC))}$, where $M_1$ is a parameter to be determined. Then  we have
$$
\begin{aligned}
    \mathcal{B}_{k,p}\leq&  MH^{2\kappa-2}k^{2\kappa}\rho(A-LC)^{\frac{2M_1\log \frac{k}{2}}{\log(1/\rho(A-LC))}}\log\frac{1}{\delta}\\
    \leq&MH^{2\kappa-2}\frac{2^{2M_1}k^{2\kappa}}{k^{2M_1}}\log\frac{1}{\delta},
\end{aligned}
$$
holds for all $k$ uniformly with probability $1-\frac{\pi^2\delta}{6}$, where the first inequality is from doubling trick. i.e., $p\ge \beta \log \frac{k}{2}$.
Hence we only need to choose $M_1> \kappa+\log H$, then there is
$$
\mathbb{P}\left\{\mathcal{B}_{k,p}\leq M\log \frac{1}{\delta},\;\; \forall k\ge T_{\text{init}}\right\}\ge 1-\frac{\pi^2\delta}{6}.
$$
For any fixed $N$, we choose $\delta = \frac{1}{N}$, then we further have
\[
\mathbb{P}\left\{\max_{k\leq N}\mathcal{B}_{k,p}\leq M\log N\right\}\ge 1-\frac{\pi^2}{6N}.
\]
With the large enough $N$, we can conclude that $\max_{k\leq N}\mathcal{B}_{k,p}\leq MH\log N$ holds almost surely.
Moreover, for the term $\mathfrak{b}_{N,p}$, note that the value of $p$ varies with the epoch index $l$. Then we divide $\mathfrak{b}_{N,p}$ apart. Similar to the previous analysis, we have  
$$
\begin{aligned}
    \mathfrak{b}_{N,p}=\sum_{l=1}^{N_E}\sum_{k=T_l}^{2T_l-2}\left\|b_{k+H, p}\right\|_{2}^{2}\leq \frac{\log(N/T_{\text{init}})}{\log2} M\log\frac{1}{\delta}
\end{aligned}
$$
holds uniformly for all $N$ with high probability $1-\frac{\pi^2\delta}{6}$.
Therefore by letting $\delta=\frac{1}{N}$, 
we can conclude that $\mathfrak{b}_{N,p}\leq M\log^2 N$ holds almost surely.
We have completed the proof that it is sufficient to guarantee the uniform boundedness of bias error only with the parameter $\beta$ chosen to be proportional to $1/\log\rho(A-LC)$ and $\kappa+\log H$.

\subsection{Proof of \cref{lemma:regression-factor}}\label{sec:proofRegression}

From \Cref{proposition:innovation}, we first have
\[r_{k+H}=e_{k+H}+\sum_{i=1}^{H-1}CA^{i-1}Le_{k+H-i}.\] Then we denote $E_{p:k}\triangleq\begin{bmatrix}
    e_p,\dots,e_k
\end{bmatrix},$ we obtain that
\[
\mathcal{R}_{k,p}=E_{p+H:k}+\sum_{i=1}^{H-1}CA^{i-1}LE_{p+H-i:k-i}
\]
With the Cauchy-Schwarz inequality, we further have
\[
\begin{aligned}
    \mathcal{E}_{k,p}\leq& H\left\|E_{p+H:k}\bar{Z}_{k-H, p}^{\tr} \bar{V}_{k-H, p}^{-\frac{1}{2}}\right\|_{2}^{2} +H\sum_{i=1}^{H-1}\left\|CA^{i-1}L\right\|_2^2\\
    &\times \left\|E_{p+H-i:k-i}\bar{Z}_{k-H, p}^{\tr} \bar{V}_{k-H, p}^{-\frac{1}{2}}\right\|_2^2.
\end{aligned}
\]
Due to the conditional independence between $e_k$ and $Z_{k-l,p}, \;\;\forall l>0$, with \cite[Theorem 3]{tsiamis9894660}, we have
\[
\begin{aligned}
    &\left\|\bar{R}^{-\frac{1}{2}}\!E_{p+H-i:k-i}\bar{Z}_{k-H, p}^{\tr} V_{k-H, p}^{-\frac{1}{2}}\right\|_2^2\\ &\qquad\qquad\qquad\qquad\leq m\log5+\log \frac{H}{\delta}+\log\frac{\det V_{k-H, p}}{\det \lambda I}.
\end{aligned}
\]
holds for all $k\ge T_{\text{init}}$ and $1\leq i\leq H-1$ uniformly with probability at least $1-\delta$, where $\bar{R}\triangleq CPC^\tr+R$ is the covariance of innovation $e_k$
Then we provide an almost sure bound for the matrix $V_{k-H,p}$.
Without loss of generality, we only need to consider the uniform bound of $V_{k,p}$ for all $k\ge T_{\text{init}}$. First note that $V_{k,p}=\lambda I+\sum_{l=p}^{k}Z_{l,p}Z_{l,p}^\tr$, then denote $\Gamma^{Z}_{k,p}\triangleq\mathbb{E}\left\{Z_{k,p}Z_{k,p}^{\tr}\right\}$, we have
$$
\begin{aligned} \left\|\Gamma_{k,p}^Z\right\|_2^2\leq&\mathbb{E}\left\{Z_{k,p}^{\tr}Z_{k,p}\right\}
    \\=&\mathrm{tr}\Big(\sum_{i=k-p+1}^{k}\mathbb{E}\left\{y_{i}y_{i}^{\tr}\right\}+\sum_{i=k-p+1}^{k+H-1}\mathbb{E}\left\{u_{i}u_{i}^{\tr}\right\}\Big)\\
    \leq&p\mathrm{tr}(R)+(p+H)n_u\\
    &\quad+\text{tr}\Big(C^{\tr}C\sum_{i=k-p+1}^{k}\sum_{l=0}^{i-1}A^l(Q+BB^\tr)(A^l)^{\tr}\Big)\\
    \leq& mp\left\|R\right\|_2+(p+H)n_u+npM\sum_{l=0}^{k-1}l^{2\kappa-2}\\
    \leq& mp\left\|R\right\|_2+(p+H)n_u+npM k^{2\kappa-1}.
\end{aligned}
$$
With \Cref{lm: Deviation}, we have
$$
\left(\Gamma_{k,p}^Z\right)^{-\frac{1}{2}}\!\! Z_{k,p}Z_{k,p}^{{\tr}}\!\!\left(\Gamma_{k,p}^Z\right)^{-\frac{1}{2}}\!\!\leq\!\! \left(\!2(mp\!+\!(p\!+\!H)n_u)\!+\!3\!\log\!\frac{k^2}{\delta}\!\right)\!I
$$
holds for all $k\ge T_{\text{init}}$ uniformly with probability $1-\frac{\pi^2 \delta}{6}$,
together with the condition $p\leq \beta \log k$, we can obtain that for each $k\leq N$, there is 
$$
\bar{Z}_{k,p}\bar{Z}_{k,p}^{\tr}=\sum_{l=p}^{k}Z_{l,p}Z_{l,p}^{\tr}\leq \left(M\beta^2 k^{2\kappa
}\log^2 k\log\frac{1}{\delta}\right)I
$$
holds uniformly with probability $1-\frac{\pi^2 \delta}{6}$. For any $N$, we further choose $\delta = \frac{1}{N}$, then we can obtain that 
\[
\bar{Z}_{k,p}\bar{Z}_{k,p}^{\tr}\leq \left(M\beta^2 N^{2\kappa
+1}\right)I, \;\;\forall k\leq N 
\]
holds uniformly with probability at least $1-\frac{\pi^2}{6N}$. Furthermore, we can obtain
\[
\begin{aligned}
    &\log \det V_{k-H,p}\leq \log\det\big(\lambda I+\bar{Z}_{k,p}\bar{Z}_{k,p}^{\tr}\big)\\
    &\qquad\qquad\qquad\leq (mp+(p+H)n_u)\log (M\beta^2 N^{2\kappa+1}),
\end{aligned}
\]
and for sufficiently large $N$, the above bound holds almost for sure. 
Then for the term $\mathcal{E}_{k,p}$, we further have that
\[
\begin{aligned}
    \mathcal{E}_{k,p}\leq& H\left(m\log 5+\log\frac{H}{\delta}+\log\frac{\det V_{k-H, p}}{\det \lambda I}\right)\\
    &\times \left\|\bar{R}\right\|_2^2\left(1+\sum_{i=1}^{H-1}\left\|C\right\|_2^2\left\|L\right\|_2^2\left\|A^{i-1}\right\|_2^2\right)\\
    \leq&MH\left(1
    +\sum_{i=1}^{H-1}i^{2\kappa-2}\right)\\
    &\times\left(\log\frac{H}{\delta}+(mp+(p+H)n_u)\log (M\beta N^{2\kappa+1})\right),\\
\end{aligned}
\]
uniformly holds for all $k\leq N$ with probability at least $1-\delta$. Similarly, for fixed $N$, with $p\leq \beta \log N$, we choose $\delta=\frac{1}{N}$, then we have the term $\log\frac{H}{\delta}=\log HN$ dominated by $p\log(M\beta N^{2\kappa+1})\leq\beta \log^2 N$. We finally have that
\[\max_{k\leq N}\mathcal{E}_{k,p}\leq MH^{2\kappa}\beta\log^2 N, 
\]
holds almost for sure.

\subsection{Proof of \cref{lemma:accumulation-factor}}\label{appendix:accumulation}
We first relax the term $\mathcal{V}_{N,p}$ as 
\[
\mathcal{V}_{N,p}\!\leq\!\! \max_{T_{\text{init}}\leq k\leq N}\left\|V_{k-H,p}^{-\frac{1}{2}}V_{k,p}^{\frac{1}{2}}\right\|_2^2\times\sum_{l=1}^{N_E}\!\sum_{k=T_l}^{2T_l-2}\!\left\|V_{k, p_l}^{-\frac{1}{2}} Z_{k,p_l}\right\|_{2}^{2},
\]
and we first consider the uniform boundedness of $\left\|V_{k-H,p}^{-\frac{1}{2}}V_{k,p}^{\frac{1}{2}}\right\|_2^2$. Note that
$$
V_{k,p}=V_{k-H,p}+\sum_{l=k-H+1}^k Z_{l,p}Z_{l,p}^\tr.
$$
Then we have
\[
\begin{aligned}
    V_{k-H,p}^{-\frac{1}{2}}V_{k,p}V_{k-H,p}^{-\frac{1}{2}}=&I+V_{k-H,p}^{-\frac{1}{2}}\Big(\sum_{l=k-H+1}^k Z_{l,p}Z_{l,p}^\tr\Big)V_{k-H,p}^{-\frac{1}{2}}.
\end{aligned}
\]
Therefore we can bound the $\left\|V_{k-H,p}^{-\frac{1}{2}}V_{k,p}^{\frac{1}{2}}\right\|_2^2$ by 
\[
\left\|V_{k-H,p}^{-\frac{1}{2}}V_{k,p}^{\frac{1}{2}}\right\|_2^2\leq 1+ \sum_{l=k-H+1}^k Z_{l,p}^\tr V_{k-H,p}^{-1}Z_{l,p} 
\]
To bound $Z_{l,p}^\tr V_{k-H,p}^{-1}Z_{l,p}$ for $l>k-H$, we need to consider the successive representation of $Z_{l,p}$ with $Z_{s,p},\; s<l$.
Consider the minimal polynomial of $A$ as
$$
A^d = a_{d-1}A^{d-1}+\dots+a_0,
$$
and the companion matrix of $A$ can be written as
\[
\mathcal{A} =
\begin{bmatrix}
0  & 0 & \cdots & 0 & a_0 \\
1 & 0 & \cdots & 0 & a_1 \\
\vdots & \vdots & \vdots & \ddots & \vdots \\
0 & 0 & \cdots & 1 & a_{d-1}
\end{bmatrix},
\]
where $d$ is the dimension of the minimal polynomial of $A$.
Then similar to \cite[Lemma 2]{tsiamis9894660}, we first derive the innovation representation of output $y_k$ as
\[
y_k=CA^{d}\hat{x}_{k-d}+\sum_{i=1}^{d}CA^{i-1}Bu_{k-i}\!+\!e_k\!+\!\sum_{i=1}^{d}CA^{i-1}Le_{k-i}.
\]
Then, with the minimal polynomial of $A$, we can derive the successive representation of the
$y_k$ as
\[
y_k=a_{d-1} y_{k-1}+\ldots+a_{0} y_{k-d}+\delta_{k}
\]
where 
\[
\begin{aligned}
   \delta_{k}&=\sum_{s=0}^d L_s e_{k-s}+ \sum_{s=1}^d K_su_{k-s},\\
   L_{s}&=-a_{d-s} I_{m}+C A^{s-1} L-\sum_{t=1}^{s-1} a_{d-s+t} C A^{t-1} L,\;L_0=I,\\
   K_s&=CA^{s-1}B-\sum_{l=1}^{s-1}a_{d-s+l}CA^{l-1}B.
\end{aligned}
\]
Then for the augmented form, further denote  $\tilde{E}_{k,p}=\left[e_{k-p+1}^\tr,\ldots,e_{k}^\tr\right]^\tr$, $\tilde{U}_{k,p}=\left[u_{k-p+1}^\tr,\ldots,u_{k}^\tr\right]^\tr$, we can rewrite $Z_{k,p}$ as
\begin{equation}\label{eq:successive}
    Z_{k,p}=a_{d-1} Z_{k-1,p}+\ldots+a_{0} Z_{k-d,p}+\delta_{k,p}.
\end{equation}
The term $\delta_{k,p}$ takes the form as $\delta_{k,p}=\left[\delta_{k,p}^{(1)\tr},\delta_{k,p}^{(2)\tr}\right]^\tr$, with
\[\delta_{k,p}^{(1)}=\sum_{s=0}^{d} \diag_p\left(L_s\right) \tilde{E}_{k-s,p}+\sum_{s=1}^{d} \diag_p\left(
K_s\right) \tilde{U}_{k-s,p}, \]
and
$
\delta_{k,p}^{(2)}=\tilde{U}_{k+H-1,p+H}-\sum_{s=0}^{d-1}a_s \tilde{U}_{k+H-1+s-d,p+H},
$
where
$\diag_p\left(L\right)=\diag\Big(\underbrace{L, \ldots, L}_p\Big)$.
Furthermore, we denote $a_{l}^{(s)}$ as the $(l+1,d)$-th element of matrix $\mathcal{A}^s$. Then substitute the successive representation of $Z_{k-1,p}$ into \cref{eq:successive}, we can obtain
\[
Z_{k,p}=a_{d-1}^{(2)} Z_{k-2,p}+\ldots+a_{0}^{(2)} Z_{k-d-1,p}+\delta_{k,p}+a_{d-1}^{(1)}\delta_{k-1,p}.
\]
By performing the above recursion for $H$ times, we can obtain
\begin{equation}\label{eq: multisuccessive}
    Z_{k,p}=\sum_{i=1}^{d}a_{i-1}^{(H)} Z_{k-H+i-d,p}+\delta_{k,p}+\sum_{j=1}^{H-1}a_{d-1}^{(j)}\delta_{k-j,p}.
\end{equation}
For each $k-H+1\leq l\leq k$, consider the term $Z_{l,p}^\tr V_{k-H,p}^{-1}Z_{l,p}$,  by substituting the successive representation \eqref{eq: multisuccessive} and applying Cauchy-Schwarz inequality, we can obtain
\begin{align}
     &Z_{l,p}^\tr V_{k-H,p}^{-1}Z_{l,p}\leq (H+1) \Big(\sum_{i=1}^{d}a_{i-1}^{(H)} Z_{l-H+i-d,p}\Big)^\tr V_{k-H,p}^{-1}\nonumber\\
&\qquad\quad\times\Big(\sum_{i=1}^{d}a_{i-1}^{(H)} Z_{l-H+i-d,p}\Big)+(H+1)\delta_{l,p}^\tr V_{k-H,p}^{-1}\delta_{l,p}\nonumber\\
&\qquad\quad+(H+1)\sum_{j=1}^{H-1}a_{d-1}^{(j)}\delta_{l-j,p}^\tr V_{k-H,p}^{-1} a_{d-1}^{(j)}\delta_{l-j,p}.
\end{align}
To provide a uniform bound for the above terms, we need to first provide a uniform bound for each $\delta_{l-j,p}$. We can verify that for fixed $d$, the norm of $L_s$ is uniformly bounded for all $s=0,\ldots, d-1$,   
Then we have the following bound for $\delta_{k,p}$ that
 \[\left\|\delta_{k,p}\right\|_{2}^2 \leq Mp\max_{k\leq N}\left\|e_{k}\right\|_{2}^2+M(p+H)\max_{k\leq N+H}\left\|u_{k}\right\|_{2}^2\]
where $M$ is only related to system parameters.
With \cref{lm: Deviation}, for fixed $N$,
we have
\[\mathbb{P}\left\{\left\|e_k\right\|_2^2\leq M\log \frac{k^2}{\delta},\;\; \forall k\ge T_{\text{init}}\right\}\ge 1-\frac{\pi^2\delta}{6}.\]
Then choose $\delta=\frac{1}{N}$, we have
\[
\max_{k\leq N}\left\|e_{k}\right\|_{2}^2\leq 3M\log N
\]
almost for sure for any fixed $N$. Moreover, due to $u_k\sim \mathcal{N}(0,I_{n_u})$, we also have
\[
\max_{k\leq N+H}\left\|u_{k}\right\|_{2}^2\leq 3M\log (N+H)
\]
almost for sure for any fixed $N$ and $H$.
Then together with $p\leq \beta \log N$ and $N\gg H$, we have
\[
\max_{k\leq N}\left\|\delta_{k,p}\right\|_{2}^2\leq M\beta \log^2 N
\]
holds almost for sure, where $H$ is eliminated as a low-order term. 

Then we reconsider the term $Z_{l,p}^\tr V_{k-H,p}^{-1}Z_{l,p},\; k-H+1\leq l\leq k$ , together with $V_{k-H,p}\ge \lambda I$, we have
\[
\begin{aligned}
a_{d-1}^{(j)}\delta_{l-j,p}^\tr V_{k-H,p}^{-1} a_{d-1}^{(j)}\delta_{l-j,p}\leq \frac{\big(a_{d-1}^{(j)}\big)^2}{\lambda}M\beta \log^2 N
\end{aligned}
\]
for each $1\leq j\leq H-1$. Moreover, from 
Woodbury Equality, there is also
$$
\begin{aligned}
    &Z_{k,p}^{\tr}\left(V_{k-1, p}+Z_{k, p} Z_{k, p}^{{\tr}}\right)^{-1}Z_{k,p}=\frac{Z_{k, p}^{{\tr}}V_{k-1,p}^{-1}Z_{k, p}}{1+Z_{k, p}^{{\tr}}V_{k-1,p}^{-1}Z_{k, p}}\leq 1.
\end{aligned}
$$
Hence we have
\[
\begin{aligned}
    \Big(\sum_{i=1}^{d}a_{i-1}^{(H)} Z_{l-H+i-d,p}\Big)^\tr V_{k-H,p}^{-1}\Big(\cdot\Big)
    \leq  d \sum_{i=1}^d \big(a_{i-1}^{(H)}\big)^2,
\end{aligned}
\]
where $(\cdot)=\sum_{i=1}^{d}a_{i-1}^{(H)} Z_{l-H+i-d,p}$ for brevity and
\[
\begin{aligned}
    Z_{l,p}^\tr V_{k-H,p}^{-1}Z_{l,p}\leq& d(H+1) \sum_{i=1}^d \big(a_{i-1}^{(H)}\big)^2+(H+1)M\beta\log^2 N\\
    &+(H+1)\sum_{j=1}^{H-1}\frac{\big(a_{d-1}^{(j)}\big)^2}{\lambda}M\beta \log^2 N.
\end{aligned}
\]
Note that the term $a_{d-1}^{(j)}$ is from matrix $\mathcal{A}^j$, and the matrix $\mathcal{A}$ shares the same minimal polynomial with $A$. Hence we have $\left\|\mathcal{A}^j\right\|_F^2\leq Mj^{2\kappa-2}$, and $\big(a_{i-1}^{(j)}\big)^2\leq Mj^{2\kappa-2}$ uniformly holds for all $i\leq d$ and $j\leq H$. Finally, we have
\[
\max_{T_{\text{init}}\leq k\leq N}\left\|V_{k-H,p}^{-\frac{1}{2}}V_{k,p}^{\frac{1}{2}}\right\|_2^2\leq MH^{2\kappa+1} \beta\log^2 N
\]
holds almost for sure, where the constant term $d \sum_{i=1}^d \big(a_{i-1}^{(H)}\big)^2 $ will also be dominated by $\log^2 N$ for large $N$.

Then we consider the term $\sum_{l=1}^{N_E}\sum_{k=T_l}^{2T_l-2}\left\|V_{k, p_l}^{-\frac{1}{2}} Z_{k,p_l}\right\|_{2}^{2}$.
For the accumulation error at the $l$-th epoch, where $T_l = 2^{l-1}T_{\text{init}}+1$, from \cite[Lemma 2]{lai1982least} (also \cite[Lemma 1]{tsiamis9894660}), we first have the following result for the accumulation error
$$
\sum_{k=T_{l}}^{2T_l-2}\left\|V_{k, p}^{-\frac{1}{2}} Z_{k,p}\right\|_{2}^{2}\leq \log\frac{\det(V_{2T_l-2,p})}{\det(V_{T_l-1,p})}.
$$
From \Cref{sec:proofRegression}, we have
\[
\log\frac{\det(V_{2T_l-2,p})}{\det(V_{T_l-1,p})}\leq \log\frac{\det(V_{2T_l-2,p})}{\det(\lambda I)}\leq M\beta\log^2 N
\]
almost for sure for each $l=1,\ldots,N_E$, then we have
\[
\sum_{l=1}^{N_E}\sum_{k=T_l}^{2T_l-2}\left\|V_{k, p_l}^{-\frac{1}{2}} Z_{k,p_l}\right\|_{2}^{2}\leq M\beta N_E\log^2 N
\]
Note that $N_E=\frac{\log N/{T_{\text{init}}}}{\log 2}$, then we have
\[
\sum_{l=1}^{N_E}\!\sum_{k=T_l}^{2T_l-2}\!\left\|V_{k, p_l}^{-\frac{1}{2}} Z_{k,p_l}\right\|_{2}^{2}\leq M\beta\log^3N
\]
and 
\[
\begin{aligned}
   \mathcal{V}_{N,p}\!\leq&\! \max_{T_{\text{init}}\leq k\leq N}\left\|V_{k-1, p}^{-\frac{1}{2}} V_{k, p}^{\frac{1}{2}} \right\|_2^2\times\sum_{l=1}^{N_E}\!\sum_{k=T_l}^{2T_l-2}\!\left\|V_{k, p_l}^{-\frac{1}{2}} Z_{k,p_l}\right\|_{2}^{2}\\
   \leq& MH^{2\kappa+1}\beta^2\log^5 N
\end{aligned}
\]
holds almost for sure for any $N$.
\subsection{Supplementary Lemma}
\begin{lemma}\label{lm: Deviation}
    For any given $\delta\in (0,1)$ and for any Gaussian random vector sequence $X_k$ satisfies $X_k\sim \mathcal{N}(0,I_{p(k)})$, where $p(k)$ is a function of $k$, we define event $\mathcal{T}_{X}$ as
    \[\mathcal{T}_{X}\triangleq\left\{\left\|X_k\right\|_{2}^2 \leq 2p(k)+3\log \frac{k^2}{\delta}, \;\; \forall k\ge 1\right\},
    \]
    then the event $\mathcal{E}_X$ holds with probability at least $1-\frac{\pi^2 \delta}{6}$.
\end{lemma}
\begin{proof}
     From \cite[Lemma 1]{laurent2000adaptive}, 
     for any Gaussian random vector $X_k\sim \mathcal{N}(0,I_{p(k)}).$ For each $k\ge 1$, we have $\mathbb{P}\left\{\left\|X_k\right\|_2^2\ge p(k)+2\sqrt{p(k)}\cdot t+2t^2\right\}\leq e^{-t^2}.$
     For each time step $k$, take $t=\sqrt{\log\frac{k^2}{\delta}}$ and with inequality $2ab\leq a^2+b^2$, then we have $\mathbb{P}\left\{\left\|X_k\right\|_2^2\leq 2n+3\log\frac{k^2}{\delta}\right\}\ge 1- \frac{\delta}{k^2}.
     $
     Then, take a union bound over all $k$, we have
     $$
     \mathbb{P}\left\{\left\|X_k\right\|_2^2\leq 2p(k)+3\log\frac{k^2}{\delta}, \forall k\ge 1\right\}\ge 1- \sum_{k=1}^{\infty}\frac{\delta}{k^2},
     $$
     Then with $\sum_{k=1}^{\infty}1/k^2=\pi^2/6$, this lemma is proved.
\end{proof}